\documentclass[letterpaper, 10 pt, conference]{ieeeconf}  

\IEEEoverridecommandlockouts                              

\usepackage{algorithm}
\usepackage{algpseudocode}
\usepackage{multirow}

\usepackage{amsmath}
\usepackage{array}        
\usepackage{booktabs}     
\usepackage{makecell}     
\usepackage{amssymb}      
\DeclareMathOperator*{\argmax}{arg\,max}

\usepackage{comment}
\usepackage{listings}
\usepackage{graphicx}
\usepackage{subcaption}
\usepackage{hyperref}

\title{\LARGE \bf Benchmarking Shortcutting Techniques for Multi-Robot-Arm Motion Planning }

\author{Philip Huang$^{1}$, Yorai Shaoul$^{1}$, and Jiaoyang Li$^{1}$
\thanks{$^{1}$Authors are with the Robotics Institute, Carnegie Mellon University, Pittsburgh, PA 15213, USA. 
        {\tt\small philiphuang@cmu.edu}}%
} 

\newcommand{\shortcut}[2]{s_{#1\rightarrow #2}}

\begin{document}

\maketitle
\thispagestyle{empty}
\pagestyle{empty}

\begin{abstract}
Generating high-quality motion plans for multiple robot arms is challenging due to the high dimensionality of the system and the potential for inter-arm collisions. Traditional motion planning methods often produce motions that are suboptimal in terms of smoothness and execution time for multi-arm systems. Post-processing via shortcutting is a common approach to improve motion quality for efficient and smooth execution. However, in multi-arm scenarios, optimizing one arm's motion must not introduce collisions with other arms. Although existing multi-arm planning works often use some form of shortcutting techniques, their exact methodology and impact on performance are often vaguely described. In this work, we present a comprehensive study quantitatively comparing existing shortcutting methods for multi-arm trajectories across diverse simulated scenarios. We carefully analyze the pros and cons of each shortcutting method and propose two simple strategies for combining these methods to achieve the best performance-runtime tradeoff. Video, code, and dataset are available at \url{https://philip-huang.github.io/mr-shortcut/}.
\end{abstract}

\section{Introduction}

Multi-robot-arm motion planning (M-RAMP) has recently seen an increased research interest, especially with the rise of bimanual manipulation systems. This trend is partly driven by the crucial role M-RAMP plays in multi-arm manipulation. The simultaneous use of multiple robot arms facilitates the automation of complex tasks infeasible with a single arm, such as collaborative robotic assembly, while also improving the efficiency of tasks typically performed by a single arm, such as pick-and-place operations. Among recent approaches to M-RAMP, robot-arm trajectories are often computed with sampling-based algorithms derived from RRT \cite{rrt} (e.g., \cite{Hartmann2023-ui}), or variants of A* search \cite{a*} (e.g., \cite{shaoul2024accelerating} and \cite{chen2022cooperativeMRAMP}).

\begin{figure}[t!]
    \centering
    \includegraphics[width=\columnwidth]{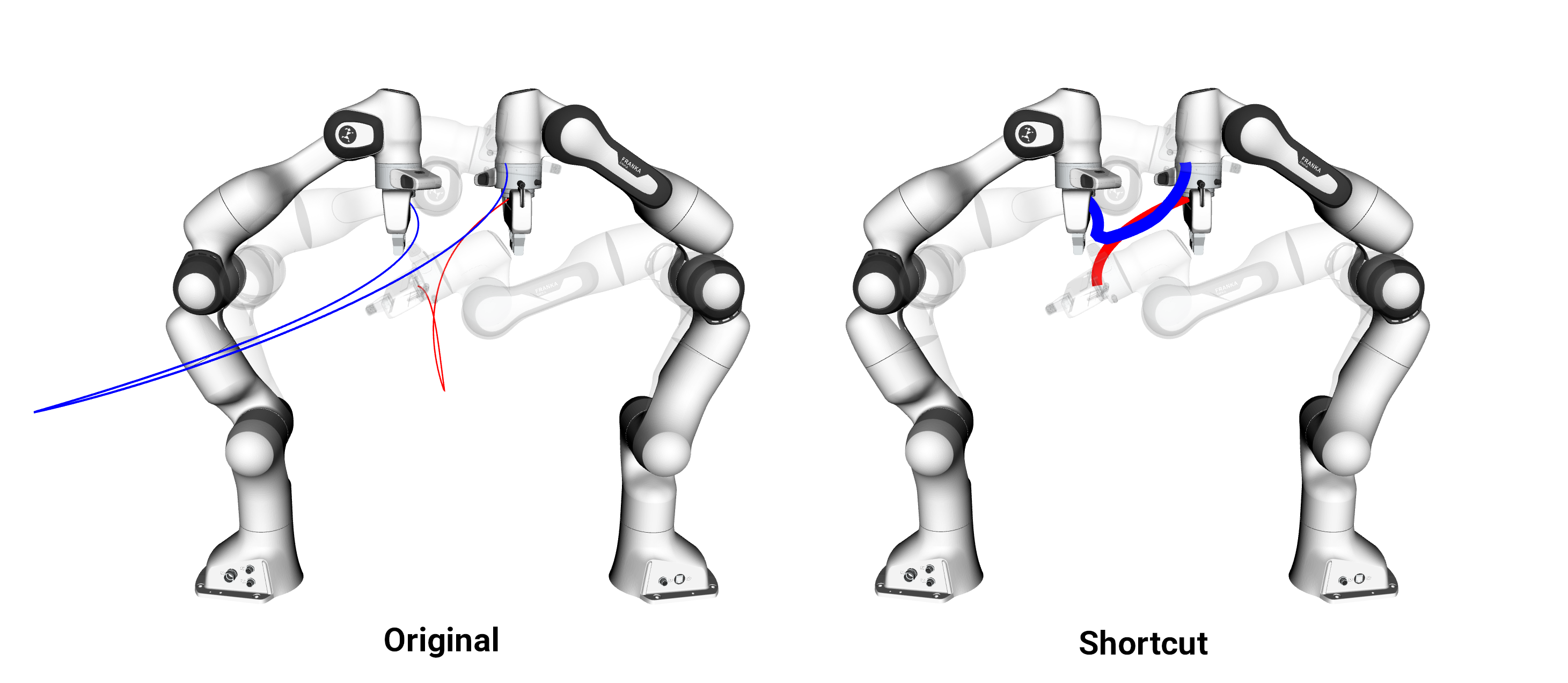}
    \caption{Dual-arm motion planned with RRT-Connect before (left) and after (right) shortcutting. Effective multi-robot shortcutting ensures smooth, collision-free robot motion.}
    \label{fig:teaser}
    \vspace{-0.5cm}
\end{figure}

A critical, yet rarely discussed, post-processing step in M-RAMP is \textit{shortcutting}. As shown in Fig. \ref{fig:teaser}, trajectories generated by sampling-based algorithms can be surprisingly long due to insufficient sampling under a tight time limit. Similarly, trajectories from search-based planner may appear choppy due to confined movement resolution on a possibly coarse graph. Shortcutting takes a trajectory as input and improves motion time, geometric length, and smoothness, making it better suited for real robot execution. In both single- and multi-arm cases, shortcutting offers significant improvements. However, in the multi-robot case, shortcutting trajectories requires more care to prevent collisions or deadlocks. Naively shortcutting individual trajectories can introduce new robot-robot collisions, highlighting the need for multi-robot shortcutting techniques. 

In stark contrast to the empirical importance of shortcutting in robotics, existing work usually only briefly mentions the multi-robot shortcutting algorithms used and often omits the impact of such shortcutting. To the best of our knowledge, multi-robot shortcutting has not been thoroughly examined in the literature before, leaving room for a better understanding of this important post-processing step.

Our contributions can be summarized as follows. First, we comprehensively survey the shortcutting methods in the existing literature that can be used for post-processing multi-robot-arm motion plans. Second, we propose two simple strategies that can adaptively combine various randomized shortcutting strategies with improved anytime performance. Third, we curate (and open-sourced) a dataset of single-task, multi-robot motion planning instances across environments with different robot configurations. 
Finally, we benchmark and analyze each method using our new dataset and show how our proposed combination strategy can combine the strengths of different algorithms.

\section{Background}


In this section, we formally introduce the multi-robot-arm motion planning problem and cover relevant work in motion planning and trajectory post-processing. 

\subsection{Multi-Robot-Arm Motion Planning}

Given $N$ robots sharing a workspace $\mathcal{W} \subset \mathbb{R}^3$, let $\mathcal{C}^i$ be the configuration space of each robot $i$ with $\text{DoF}^i$ degrees of freedom (DoF). For robot arms with ${\text{DoF}^i}$ joints, a configuration $C^i \in \mathcal{C}^i \subseteq \mathbb{R}^{\text{DoF}^i}$ defines a selection of joint angles. Given a pair of start and goal configurations for each robot $i$, M-RAMP focuses on finding a set of collision-free \textit{trajectories} $\tau := \left\{\tau^1, \cdots \tau^N\right\}$, one for each robot. We represent a single-robot trajectory with a sequence of $H^i$ timestamped configurations
$\tau^i = \{\tau^i_n \}_{n=1}^{H^i} = \{(C_n^i, t_n^i)\}_{j=1}^{H^i}$, with $C_n^i$ being the configuration of robot $i$ at time $t_n^i$. In this paper, each step ($C_n^i, t_n^i$) to ($C_{n+1}^i, t_{n+1}^i$) is over a constant time interval $\Delta t$ and spatially defined as a linear interpolation between the two configurations, and $t_n^i$ at each step $n$ is the same for all robots, i.e., $t_n^i = t_n^j \; \forall \; i \neq j$. M-RAMP seeks low-cost trajectories, minimizing travel time or cumulative motion, while satisfying collision constraints (i.e., no robot-robot or robot-world collisions) and velocity constraints (i.e., $d(C^i_j, C^i_{j+1}) \leq \Delta t \cdot v_{\max}^i$ for each step, with $d(\cdot, \cdot)$ being an $L_1$ distance (radians) and $v_{\max}^i$ a cumulative speed limit).

Researchers have proposed various approaches to multi-robot motion planning and applied them to M-RAMP, including directly using single-robot planners, such as RRT-Connect \cite{Kuffner2000-hn}, BIT* \cite{Gammell2020-bd}, or graphs of convex sets \cite{Marcucci2023-fo}, in the composite configuration space $\mathcal{C}:= \mathcal{C}^1 \times \cdots \times \mathcal{C}^N$. Other approaches more tailored towards M-RAMP include building composite roadmaps from the Cartesian product of individual roadmaps \cite{Gharbi2009-rv}, coordinating the speed of individually planned motions \cite{Bien1992-ze, Zhang_undated-hg}, or using Conflict-Based-Search (CBS) \cite{sharon2015conflict} on roadmaps \cite{solis2021representation} or configuration space lattices \cite{shaoul2024accelerating}. Researchers have also studied more long-horizon manipulation tasks with multiple robot arms, such as object pick and place \cite{Harada2014-zp, Gao2022-iu, Gao2024-bl} and object rearrangement and assembly \cite{ Shome2021-ka, Hartmann2021-kf, huang2025apexmr, chen2022cooperativeMRAMP, Pan2021-ar}.
Many of the multi-robot motion plans also use post-processing to improve their motion plans obtained.

\subsection{Trajectory Post-Processing}

Algorithms used to post-process single-robot trajectories can be classified as shortcutting, gradient-based, or retiming. Gradient-based trajectory optimization methods such as CHOMP \cite{Ratliff2009-yp} and KOMO \cite{Toussaint2014-eg} can be used to post-process a planned trajectory, but do not guarantee that the solution remains collision-free. Methods like Toppra \cite{Pham2018-be} tune the velocity of a trajectory to reduce execution time, not considering any further gains that could be made by altering the robot's path. Shortcutting iteratively or randomly finds a collision-free linear interpolation between points on a trajectory \cite{Geraerts2007-mq, choset2005shortcutting}, guaranteeing that the solutions remain collision-free, and reduces execution time.

In multi-robot systems, the main post-processing objective is to reduce the \textit{makespan} of the set of trajectories $\tau$. That is, finding a new collision-free trajectory $\tau'$ between the same starts and goals of $\tau$ that minimizes $\max_i t_{H^i}^i$.
\footnote{In multi-robot settings, optimizing for time may be more appropriate than optimizing for path length. When multiple robots are coordinated, one robot may wait in place to avoid collisions with others, yielding a spatially short path but a long execution time.} 
A challenge specific to multi-robot motion post-processing is that altering the trajectory of one robot could affect others through new robot-robot collisions. \cite{Gao2024-bl, Shome2021-ka} circumvent this challenge by retiming the trajectories of all robots jointly in a centralized fashion, reducing the makespan but with limited improvement. Other approaches like \cite{shaoul2024accelerating} use iterative shortcutting to smooth each robot's motion while treating other robots as obstacles. \cite{Hartmann2021-kf} uses both randomized shortcutting and single-robot trajectory optimization to post-process each robot's motion segment individually during planning. \cite{Okumura2022-mw} and \cite{huang2025apexmr} randomly find shortcuts on a Temporal Plan Graph (TPG) \cite{Honig2016-ts} representation of multi-robot trajectories.

\section{Multi-Robot Shortcutting Algorithms}
\vspace{-0.1cm}
In this section, we present three common shortcutting methods for single robots, followed by four different methods that generalize these to multi-robot setups. We then compare the advantages and limitations with examples and analysis. 

\vspace{-0.1cm}
\subsection{Single-Robot Shortcutting}
\vspace{-0.1cm}
Given a single-robot trajectory $\tau = \{\tau_j \}_{j=1}^{H}$, shortcutting methods find a pair of configurations $C_m$ and $C_n$ on $\tau$ and try to connect them with a shortcut $\shortcut{m}{n}$, computed via linear interpolation between $C_m$ and $C_n$. $\shortcut{m}{n}$ is valid if it is shorter than the original trajectory segment $\{\tau_j\}_{j=m}^n$ and collision-free with the environment. To maintain consistency, $\shortcut{m}{n}$ is discretized with the same granularity as $\tau$.
These methods iteratively test different pairs of $C_m$ and $C_n$ and replace $\{\tau_j\}_{j=m}^n$ with $\shortcut{m}{n}$ in $\tau$ when $\shortcut{m}{n}$ is valid. 
There are three common ways of enumerating these pairs: 
\begin{enumerate}
    \item \textbf{Randomized}~\cite{Sekhavat1998-rs} samples the pairs uniformly random.
    \item \textbf{Forward Loop}~\cite{Hsu1999-in, Chen1998-sy} iterates over all pairs forward, using an outer loop incrementing $m$ from $1$ to $H-1$ and an inner loop incrementing $n$ from $m+2$ to $H$. It checks if $\shortcut{m}{n}$ is a valid and accepts it if so. 
    \item \textbf{Backward Loop}~\cite{Hsu1999-in, Chen1998-sy} modifies the forward loop method by decrementing $n$ from $H$ to $m+1$ backwards in the inner loop. 
\end{enumerate}
\vspace{-0.2cm}

\subsection{Multi-Robot Shortcutting}
\vspace{-0.1cm}
When dealing with multiple robots, single-robot shortcutting methods cannot be directly applied to each robot independently, as shortening the trajectory segment of one robot may lead to collisions with others. To address this challenge, four distinct multi-robot shortcutting methods are used in practice. Each method employs a different strategy to ensure multi-robot trajectories remain collision-free after shortcutting. Importantly, these strategies are orthogonal to the configuration pair selection strategies in single-robot shortcutting. This modularity means that any of the four multi-robot methods can be combined with any of the three single-robot shortcutting approaches mentioned above.

1) \textbf{Composite-space Shortcutting} \cite{Svestka1998-cb} treats all robots as a single agent by combining all degrees of freedom from all robots into a single composite space $\mathcal{C}$.
A shortcut $\shortcut{m}{n}$ comprises single-robot shortcuts that connect ${C_m^i}$ and ${C_n^i}$ for each robot $i$. It discretizes $\shortcut{m}{n}$ into a sequence of timestamped configurations in the composite space such that no robot exceeds its speed limit between two successive configurations. 
The shortcut is collision-free if all of its interpolated configurations are collision-free. If accepted, each robot's trajectory is shortened by the same amount, preventing inter-robot collisions.

2) \textbf{Prioritized Shortcutting}
finds a shortcut for one robot's trajectory at a time, treating other robots as dynamic obstacles that it must avoid. 
It randomly selects a robot $i$ (or cycles through the robots in a round-robin fashion) and tries a shortcut between $C_m^i$ and $C_n^i$. The shortcut $\shortcut{m}{n}$ is discretized into timestamped configurations, assuming that robot $i$ travels at maximum speed. Since $\shortcut{m}{n}$ reduces the time for robot $i$ to reach $C_n^i$, the timesteps of the remaining trajectory $\{\tau^i_j\}_{j=n+1}^{H^i}$ decrease by the same amount as $t_n$ has been reduced. To accept a shortcut, it checks both if $\shortcut{m}{n}$ is collision-free and if the remaining trajectory of robot $i$ with shortened timesteps collides with other robots.

3) \textbf{Path Shortcutting}~\cite{Hartmann2021-kf, shaoul2024accelerating} improves prioritized shortcut by eliminating collision checking for the trajectory after $C_n^i$. The key idea is to keep all timesteps unchanged and modify only the configurations between $C_m^i$ and $C_n^i$. $\shortcut{m}{n}$ is uniformly discretized into the same number of configurations as in the original trajectory segment, indicating that the robot moves along a shorter trajectory but at a slower speed. After shortcutting, the method retimes the trajectory in the composite space to minimize the makespan, resetting the timesteps so that the fastest robot in each interval is traveling at its maximum velocity. Specifically, we set $t_n=t_{n-1}+\max_i(d(C^i_n, C^i_{n-1})/v_{max}^i)$ iteratively from $n=1$. 

4) \textbf{TPG Shortcutting}~\cite{Okumura2022-mw, huang2025apexmr} is motivated by Temporal Plan Graphs (TPG)~\cite{Honig2016-ts}, an execution framework from the Multi-Agent Path Finding (MAPF) community. TPG is a directed graph that captures the relative execution order of given multi-robot trajectories. Each node $v_m^i$ represents a discretized configuration $C_m^i$, while directed edges $(v_m^i, v_n^j)$ describe precedence constraints. Edges from $v_m^i$ to $v_{m+1}^i$ for all $m$ and $i$ preserve each robot's configuration order. For every pair of colliding configurations $C_m^i$ and $C_n^j$ ($i \neq j$ and $t_m < t_n$), an edge from $v_{m+1}^i$ to $v_n^j$ ensures robot $i$ needs to travel through $C_m^i$ and reach $C_{m+1}^i$ before robot $j$ moves to $C_n^j$. The TPG allows robots to move asynchronously as long as the precedence constraints are satisfied. Critically, shortening a robot $i$'s trajectory segment requires no collision checking with trajectory segments of other robots that cannot be executed simultaneously due to precedence constraints. 

Therefore, recently, Okumura et al. \cite{Okumura2022-mw} proposed using TPG to shortcut multi-robot trajectories. For a shortcut $\shortcut{m}{n}$ of robot $i$, it discretizes $\shortcut{m}{n}$ assuming robot $i$ travels at maximum speed. For every discretized configuration in $\shortcut{m}{n}$, it checks collisions with any other nodes from other robots that are neither a predecessor of $v_m^i$ or a successor of $v_{n}^i$\footnote{We can skip collision checking with these nodes because they must have precedence constraints with the node being checked, so the corresponding configurations will never be reached by the robots simutenously.} and accepts the shortcut if no collisions are found. 

\vspace{-0.15cm}
\subsection{Comparison}
\vspace{-0.1cm}
Below, we analyze the strengths and weaknesses of these single and multi-robot shortcutting methods by examining their computational speed and effectiveness.


\subsubsection{Single-Robot Shortcutting: Randomized vs Iterative}
\label{sec:single-robot-comp}
We first analyze whether the two endpoints of a shortcut $C_m$ and $C_n$ should be chosen iteratively or randomly. Forward iteration finds more shortcuts but improves less per valid shortcut. This is because candidate shortcuts are more likely to originate from nearby configurations with less potential for makespan reduction and requiring additional shortcuts, along with more collision checks, to converge. Searching for shortcuts iteratively backward is more greedy. The shortcut can yield more significant reductions on average, but a long shortcut between two distant configurations is much less likely to be collision-free. Randomized sampling has a uniform probability of selecting each possible shortcut, thus exploring all configurations on a robot trajectory much more efficiently.  Empirically, we also find that randomized sampling reduces the path length and makespan more efficiently.

\begin{figure}[t!]
    \centering
    \includegraphics[width=\linewidth]{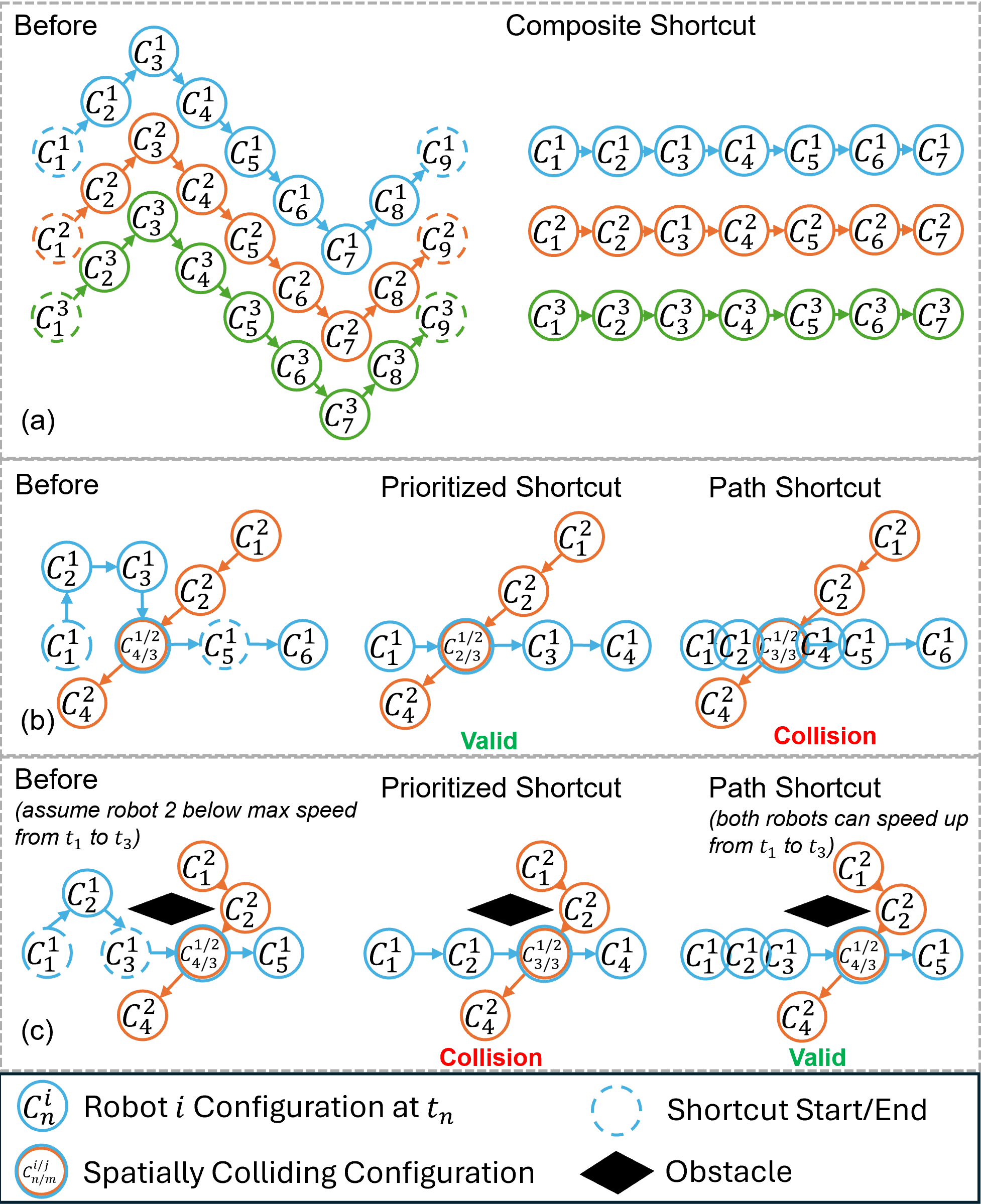}
    \caption{Toy examples in 2D that highlight the strengths and weaknesses of composite, prioritized, and path shortcutting. In (a), composite shortcutting is more efficient than prioritized and path shortcutting. In (b), only prioritized shortcutting can find a collision-free shortcut. In (c), only path shortcutting can find a collision-free shortcut and reduce the path length for robot 1. Since we assume robot 2 was moving below max speed from $t_1$ to $t_3$, both robots can move faster from $t_1$ to $t_3$ after retiming and reduce makespan.}
    \label{fig:shortcut_exp}
    \vspace{-0.8cm}
\end{figure}

\subsubsection{Multi-Robot Shortcutting: Computational Speed}
The speed of a shortcutting method is largely dominated by the amount of time spent on collision checking between robot configurations. 
Composite shortcuts modify all robots' configurations simultaneously, whereas prioritized or path shortcuts modify one robot's path at a time. This implies that composite shortcutting requires fewer shortcuts, and thus, fewer collision checks.
Compared to prioritized shortcutting, path shortcutting is faster because it does not check for configurations after the shortcut endpoint $C_n^i$.

 
TPG shortcutting, however, is computationally expensive, requiring more collision checks, especially in graphs with few inter-robot edge dependencies. The worst-case complexity of collision checking each candidate shortcut grows quadratically with the number of robots and trajectory length. Also, the shortcut must be collision-free with more spatial configurations. For instance, if there is no inter-robot edge in the TPG, any shortcut must be collision-free with the entire path of all other robots to be valid, increasing runtime complexity and degrading final trajectory quality.

\subsubsection{Multi-Robot Shortcutting: Effectiveness}
We compare how different multi-robot shortcutting methods perform using three illustrative 2D multi-robot examples in Fig. \ref{fig:shortcut_exp}.

In Fig. \ref{fig:shortcut_exp}(a), composite shortcutting can produce valid shortcuts for any choice of two indices ($n \rightarrow m$), efficiently reducing all robots' path lengths simultaneously for makespan reduction. In contrast, the best possible prioritized and path shortcuts are either $C^1_3 \rightarrow C^1_9$ or $C^3_1 \rightarrow C^3_7$ in the first iteration. No shortcuts for robot 2 are valid until the path for robot 1 or 3 changes, requiring both methods to attempt many more shortcuts to reach the optimal solution.

In Fig. \ref{fig:shortcut_exp}(b), composite shortcutting fails because robot 2 is already traveling at max speed on the shortest path. By definition, it must interpolate all robots' configurations at the same time while no robot exceeds its speed limit. Prioritized shortcutting allows robot 1 to pass through the spatially conflicting configuration earlier than robot 2 and reach the old $C_5^1$ two steps earlier. For path shortcutting, robot 1 would still need four steps to reach $C_5^1$, which would result in a collision with robot 2 at $t_3$. Prioritized shortcutting allows robots to switch the order when passing a spatially conflicting configuration, while path shortcutting does not.

In Fig. \ref{fig:shortcut_exp}(c), composite shortcutting fails due to the static obstacle blocking robot 2's shorter paths (e.g., a shortcut from $C_1^2 \rightarrow C_3^2$). Prioritized shortcutting is infeasible because robot 1 would arrive one step earlier at the spatially conflicting $C_3^2$ and collide with robot 2. A path shortcut is valid because robot 2 still passes through the conflicting pose first and thus avoids collisions. Since we assume that robot 2 is below its speed limit from $t_1$ to $t_3$, the makespan can be reduced by retiming the trajectory and speeding up both robots after the path shortcutting. If robot 2 is at its maximum speed limit, then the path shortcut only reduces the path length but does not reduce the makespan.

Overall, the best shortcutting method is trajectory-dependent. Between path and prioritized shortcutting, one method may identify valid shortcuts while the other fails. Composite shortcutting uses the fewest shortcuts. We excluded the TPG shortcut from these examples due to its high computational cost, as its performance was consistently inferior to the other three methods in our experiments.


\vspace{-0.1cm}
\section{Multi-Strategy Methods}
\vspace{-0.1cm}
Since no single multi-robot shortcutting method performs best universally, we can mix different multi-robot shortcutting methods across iterations. Our multi-strategy methods combine composite, prioritized, and path shortcutting with randomized shortcutting. In each iteration, we select one shortcutting approach and use it to select one shortcut (e.g., prioritized shortcutting selects one robot's path to shortcut). If successful, the shortcut updates the corresponding portion of the trajectory and repeats until the time limit is reached. We use randomized single-robot shortcutting because it performs best empirically, and we exclude TPG shortcutting due to its time-consuming construction and maintenance.
 

We introduce two simple multi-strategy methods, both requiring no pre-processing and minimal runtime overhead.

\subsubsection{Round Robin (RR)} This simple strategy cycles through the shortcutting methods until an iteration limit. RR has also been used to select heuristics in suboptimal search \cite{Fickert2022-fo}.

\subsubsection{Thompson Sampling}
Dynamic Thompson Sampling (DTS) \cite{Gupta2011-gk} models the shortcutting algorithm selection as a multi-arm bandit problem with a shifting underlying distribution. The runtime and shortcut improvement of each method determines the reward. We adapt the DTS from the multi-heuristic search in \cite{Phillips2015-ar}. 
Mathematically, we initialize parameters $\alpha_i$ and $\beta_i$ of the beta distribution used to model the probability of success for each shortcutting strategy $i$. 
At each iteration, we sample $\theta_i$ from beta distribution $Beta(\alpha_i, \beta_i)$ for each strategy $i$, then select the strategy $i^* = \argmax_i \theta_i$ to attempt a shortcut.
If the shortcut is accepted, we compute a reward as $r = d + \gamma_t \max(0, 1-t/ \sigma)$, where $d$ is the relative path length reduction, $t$ is the runtime for evaluating shortcuts, $\sigma$ is the normalizing time scale, and $\gamma_t$ is the relative weight to reward a shorter runtime. The reward prioritizes greater path length reduction in less time.
If accepted, we update $\alpha_{i^*} \gets \alpha_{i^*} + \gamma_{\alpha} \cdot r$, where $\gamma_{\alpha}$ is the update weight for $\alpha$.
If the shortcut is rejected, we update $\beta_{i^*} \gets \beta_{i^*} + \gamma_{\beta}$. 
The sum of $\alpha_{i^*}$ and $\beta_{i^*}$ is normalized to the normalization constant $Z$ if it exceeds $Z$.



\vspace{-0.1cm}
\section{Experiments}
To assess the efficacy of each shortcutting method, we generated 1,034 multi-robot-arm trajectories with two motion planning algorithms in 6 environments and shortcutted them with each of our benchmarked methods. Specifically, we designed the benchmark to answer the following questions: 1) How do existing shortcutting methods compare across environments with different complexity? 2) Does the proposed algorithmic selector combine the strengths of each method? 

\vspace{-0.1cm}
\subsection{Evaluation environment}
\begin{figure}[t]
    \centering
    \includegraphics[width=\columnwidth]{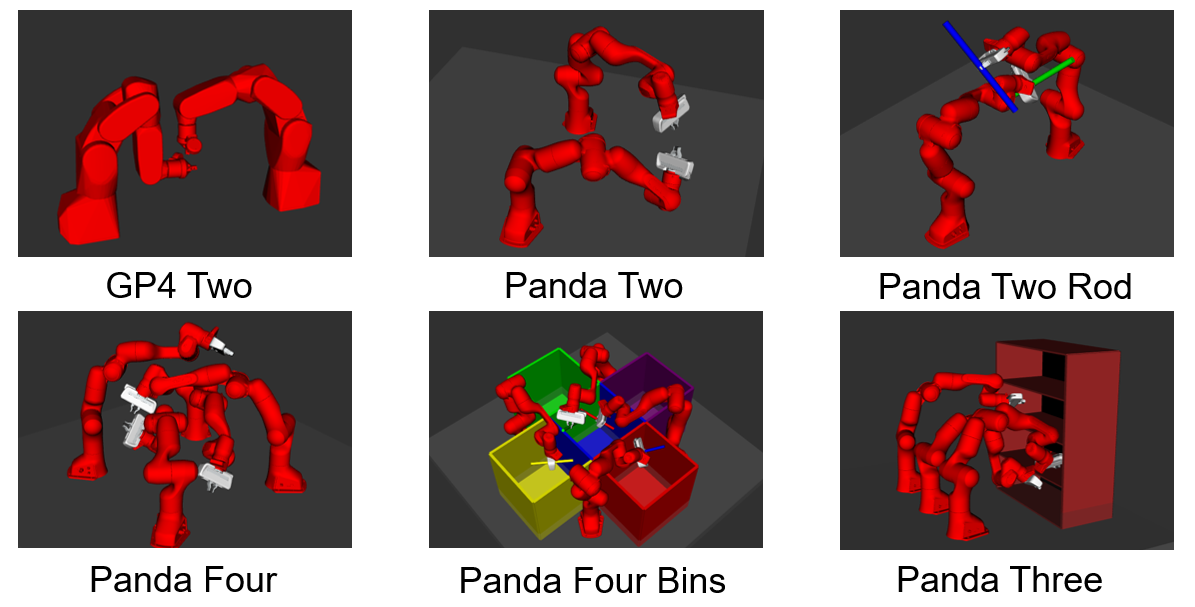}
    \caption{Visualization of the six multi-robot environments. }
    \label{fig:envs}
    \vspace{-0.6cm}
\end{figure}

As shown in Fig. \ref{fig:envs} and Table \ref{tab:robot_envs}, we create 6 evaluation environments in \textit{MoveIt!}  \cite{Coleman2014-dq}, a popular software package for robot arm motion planning. We design problem instances with varying difficulties by adjusting the proximity among robots at start and goal configurations, attaching objects to their end effector, and changing the environment. Our setup includes two robot types: 6-DoF Yaskawa GP4 and 7-DoF Franka Panda. To make the evaluation more realistic, we attach long rods to the robot's end effector in the Panda Two Rod and Panda Four Bins, significantly increasing planning complexity due to more collision avoidance requirements. 

\vspace{-0.1cm}
\subsection{Evaluation Metrics}
Motion planning literature has a variety of different optimization objectives, often tailored for specific objectives such as path length, clearance, motion cost, planning time, or higher-order derivatives such as jerk. Performance can vary depending on the metric chosen, but we focus on the kinematic path quality in multi-robot systems as this is the most fundamental aspect to robot motion planning. The quality of a kinematic trajectory includes the execution time (e.g., makespan), physical path length, and smoothness. The efficiency of the shortcutting algorithm is also important. Thus, we choose the following metrics to evaluate the performance of different shortcutting methods.
\begin{itemize}
    \item \textbf{Makespan}(s) $\max_i t_{H^i}^i$: maximal time among robots.
    \item \textbf{Path Lengths}(rad) $\sum_{i=1}^N \sum_{j=2}^{H_i} d( C_{j-1}^i, C_{j}^i )$: The total length of the paths, excluding any wait actions.
    \item \textbf{Directional Consistency} $\sum_{i=1}^N \sum_{j=2}^{H_j} (1 - \frac{v_{j-1}^i \cdot v_j^i }{||v_{j-1}^i|| \cdot ||v_j^i||})$, where $v_j^i = C_{j+1}^i - C_{j}^i$. It measures the number and degree of ``kinks" in the path with cosine similarity. 
    \item \textbf{Runtime}: The time used by the shortcutting algorithm.
\end{itemize}

We also report additional statistics, such as the number of evaluated and valid shortcuts, to highlight the difference in behavior and characteristics of different algorithms.

\subsection{Motion Planning Algorithms} In our experiments, we generated multi-robot trajectories using two algorithms, each representing a different approach to motion planning.
\subsubsection{Sampling-Based Motion Planner} RRT-Connect \cite{Kuffner2000-hn}, a popular and effective motion planner for robotic manipulators, offers quick computation time but can produce occasional poor motion quality (see Fig. \ref{fig:teaser}). Using RRT-Connect in multi-arm settings is straightforward. In multi-arm settings, we treat all arms as a single, higher-dimensional meta-robot, with its configuration space as the Cartesian product of individual arm configuration spaces. RRT-Connect samples this composite configuration space to find a trajectory between the composite start and goal configurations. We use the OMPL \cite{Sucan2012-dx} implementation of RRT-Connect.

\subsubsection{Search-Based Motion Planner} xECBS \cite{shaoul2024accelerating}, a state-of-the-art motion planner for multi-arm settings, builds on the CBS \cite{sharon2015conflict} algorithm. It begins by computing single-robot trajectories for all arms independently, usually using A* \cite{hart1968A*}. Then, CBS incrementally resolves robot-robot collisions by imposing constraints and replanning. To cast multi-arm motion planning as a MAPF problem, CBS embeds the configuration space of each arm into a graph, where vertices are configurations and edges are kinematically valid transitions between configurations. 
Our implementation is similar to the one outlined in \cite{shaoul2024accelerating}.

As shown in Table \ref{tab:robot_envs}, RRT-Connect trajectories are often longer with unnecessary motions. CBS can find trajectories with shorter path length but are often less smooth, leaving room for improvement.


\begin{table}[t]
    \centering
    \small
    \setlength{\tabcolsep}{2pt} 
    \captionof{table}{Summary of multi-arm environments characteristic and solution metrics for sampling-based RRT and search-based CBS. Average path length (PL), makespan (MK), and directional consistency (DC) are reported. }
    \begin{tabular}{@{\hskip 2pt}c@{\hskip 2pt}c@{\hskip 2pt}c@{\hskip 2pt}c@{\hskip 2pt}c@{\hskip 2pt}c@{\hskip 2pt}c@{\hskip 2pt}c@{\hskip 2pt}c@{}}
    \toprule
    \multirow{2}{*}{\makecell{Environment}} & \multirow{2}{*}{\makecell{Static\\Obstacles}} & \multirow{2}{*}{\makecell{Attached\\Objects}} & \multicolumn{2}{c}{PL} & \multicolumn{2}{c}{MK} & \multicolumn{2}{c}{DC} \\ 
        \cmidrule(lr){4-5} \cmidrule(lr){6-7} \cmidrule(lr){8-9}
    & & & RRT & CBS & RRT & CBS & RRT & CBS \\ \midrule
    Dual GP4        & $\times$       & $\times$       &    8.6     &    7.4     &     5.0    &    4.9     &   0.00      &   0.21      \\
    Panda Two       & $\times$       & $\times$       &    15.9     &    7.7     &    9.3     &    5.0     &    0.01     &     0.16    \\
    \multirow{2}{*}{\makecell[c]{Panda Two\\Rod}}   & \multirow{2}{*}{$\times$}       & \multirow{2}{*}{\checkmark}     &    \multirow{2}{*}{20.2}     &     \multirow{2}{*}{11.4}    & \multirow{2}{*}{11.7} & \multirow{2}{*}{7.5} & \multirow{2}{*}{0.01} &    \multirow{2}{*}{0.17}     \\
    &&&&& \\
    Panda Three     & \checkmark     & $\times$       &    29.9    &    23.7     &    12.4     &    11.3     &    0.00     &    0.17     \\
    Panda Four      & $\times$       & $\times$       &    36.2     &    18.2     &     11.4    &     7.0    &    0.00     &    0.16     \\
    \multirow{2}{*}{\makecell[c]{Panda Four\\Bins}} & \multirow{2}{*}{\checkmark}     & \multirow{2}{*}{\checkmark}     &   \multirow{2}{*}{46.2}    &    \multirow{2}{*}{17.5}     &     \multirow{2}{*}{14.4}     &     \multirow{2}{*}{7.1}     &     \multirow{2}{*}{0.01}     &     \multirow{2}{*}{0.14}     \\
    &&&&& \\
    \bottomrule
    \end{tabular}
    \label{tab:robot_envs}
    \vspace{-0.4cm}
\end{table}

\begin{figure*}[h!]
    \centering
    \includegraphics[width=\textwidth]{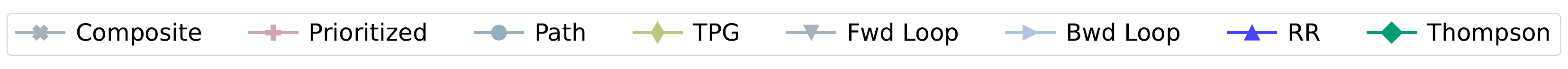}
     
    \begin{subfigure}[b]{0.49\textwidth}
        \centering
        \includegraphics[width=\columnwidth]{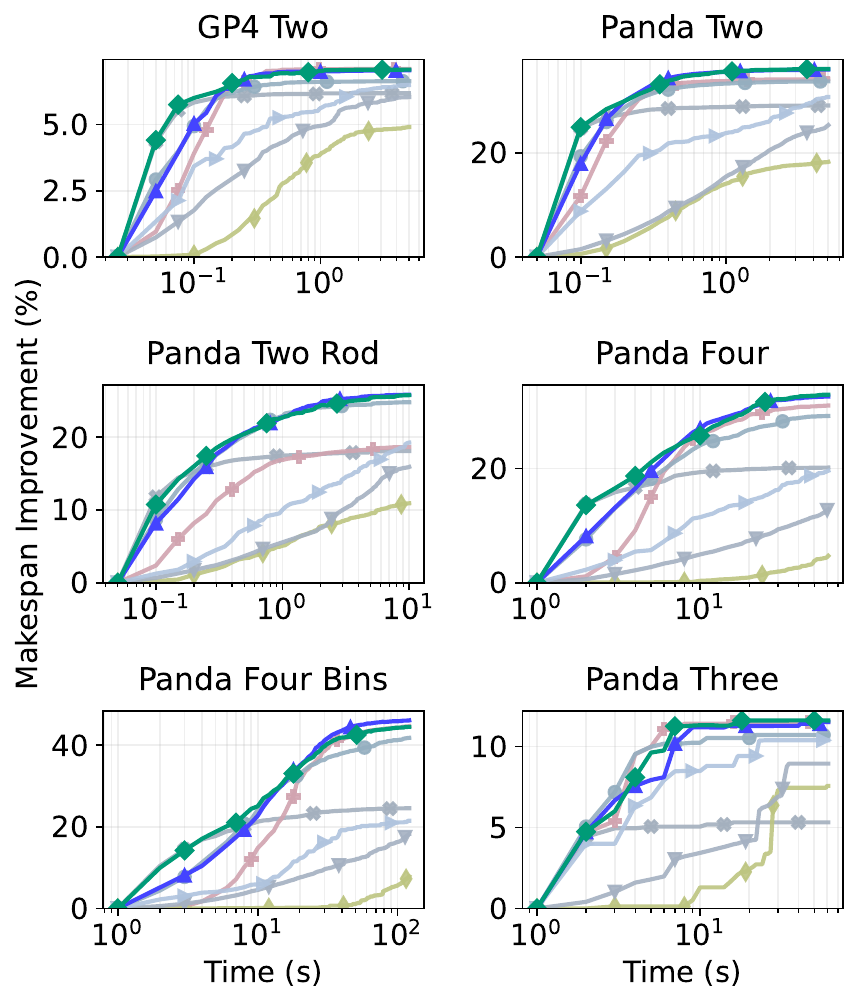}
        \caption{RRT-generated trajectories}
        \label{fig:makespan_rrt}
    \end{subfigure}
    \hfill
    \begin{subfigure}[b]{0.49\textwidth}
        \centering
        \includegraphics[width=\columnwidth]{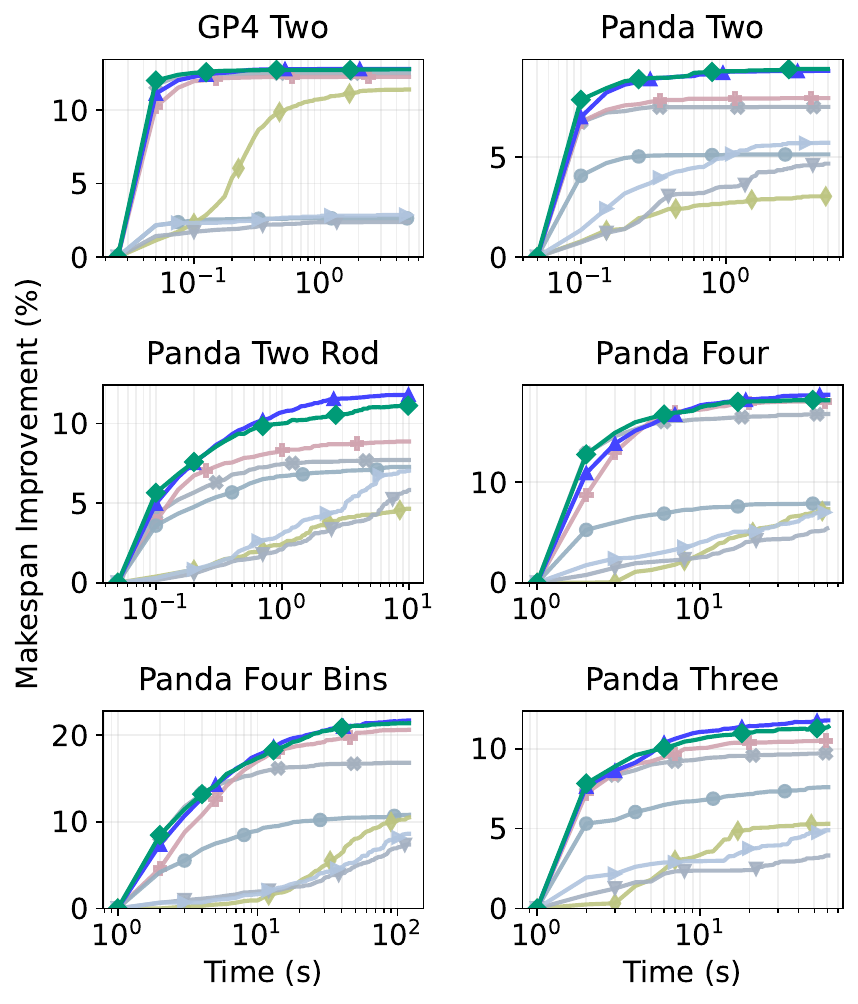}
        \caption{CBS-generated trajectories}
        \label{fig:makespan_cbs}
    \end{subfigure}
    
    \caption{Comparison of average makespan improvement over time across different environments for benchmarked shortcut methods on both RRT- and CBS-generated trajectories. The two multi-strategy methods (RR and Thompson) outperform the others. Composite, Prioritized, Path, and TPG combine the respective multi-robot shortcutting strategy with randomized single-agent shortcutting. Fwd Loop (Forward Loop) and Bwd Loop (Backward Loop) combine the respective single-robot shortcutting strategy with path shortcutting to generalize to a multi-robot setting. }
    \label{fig:makespan_comparison}
    \vspace{-0.5cm}
\end{figure*}

\subsection{Implementation Detail}
The runtime limit for the shortcutting algorithm is 5 seconds for the GP4 Two, 10 seconds for both Panda Two, 120 seconds for the Panda Four Bins, and 60 seconds for other Panda arm environments. The runtimes are selected based on our estimates of the time required for randomized composite, prioritized, and path shortcutting methods to converge. Note that because shortcutting algorithms are anytime, it is unnecessary to run them for a long time in practice.

If the initial planned trajectory uses non-uniform timesteps, we retime the trajectory to a uniform step size $\Delta t$ by linear interpolation. Collisions are checked between the geometry of each robot (including any attached objects) with other robots and the environment at each discretized timestep. Thus, picking a suitable $\Delta t$ is important as it directly affects the resolution and complexity of collision checking.
We found that setting the uniform discretization step to be $\Delta t = 0.1s $ and the maximum $L_1$ velocity for each robot arm to be $1 rad/s$ achieves a good balance between runtime and accuracy. 
Detailed parameters of Thompson sampling are available in the appendix.
We implemented all shortcutting algorithms in C++ and conducted our experiment on a server with an AMD Threadripper 3990X CPU.

\subsection{Results}
We summarize all performance metrics and the statistics of all benchmarked shortcutting methods in Fig. \ref{fig:makespan_comparison} and Table \ref{tab:agg_stat}. 
We report the results separately for robot trajectories planned by RRT and CBS because the planned trajectories have very different characteristics. Instead of evaluating all possible ways to generalize single-robot shortcutting to multi-robot, we evaluate two groups of representative combinations. The first group - Composite, Prioritized, Path, and TPG all use the randomized single-robot method with respective multi-robot methods. The second group - Fwd Loop (Forward Loop) and Bwd Loop (Backward Loop), uses path shortcutting with the respective single-robot iteration method. We report multi-strategy methods - RR and Thompson - in a third group.

Across all settings, multi-strategy shortcutting methods outperform all other single-strategy methods regarding the final execution time, path length, and smoothness. The best shortcutting algorithm significantly improves the makespan of CBS-generated trajectories by around 14\% and RRT-generated trajectories by 25\%, which indicates the importance of shortcutting. Our earlier analysis of the different shortcutting strategies aligns with the empirical evaluations, as discussed below.

\begin{table*}[htbp!]
    \centering
    \caption{Average of selected metrics (\% and standard deviation) measured across all trajectories from all six environments. The \textbf{bold} entry indicates the best-performing algorithm in the main metrics.}
    \label{tab:agg_stat}
    \begin{tabular}{ll|ccc!{\vrule width 1pt}ccc}
\toprule
Planner & Method & \makecell{Makespan\\Improvement\\(\%)} & \makecell{Path Length\\Improvement\\(\%)} & \makecell{Directional\\Consistency\\Improvement (\%)} & \makecell{\# Candidate\\Shortcuts} & \makecell{\# Valid\\Shortcuts} & \makecell{Makespan\\Improvement per\\ Valid Shortcuts (s)} \\
\midrule
\multirow{8}{*}{CBS} & Composite & 12.1 $\pm$ 7.7 & 2.4 $\pm$ 3.2 & 58.7 $\pm$ 30.6 & 2.0k $\pm$ 3.0k & 3 $\pm$ 3 & 0.260 $\pm$ 0.216 \\
& Prioritized & 13.1 $\pm$ 8.8 & 3.0 $\pm$ 3.9 & 54.4 $\pm$ 27.8 & 2.6k $\pm$ 3.9k & 8 $\pm$ 7 & 0.137 $\pm$ 0.122 \\
& Path & 6.2 $\pm$ 6.2 & 3.9 $\pm$ 4.7 & 34.8 $\pm$ 28.3 & 2.3k $\pm$ 3.3k & 11 $\pm$ 15 & 0.087 $\pm$ 0.116 \\
& TPG & 7.8 $\pm$ 8.4 & 1.6 $\pm$ 2.6 & 37.4 $\pm$ 33.8 & 1.6k $\pm$ 3.6k & 11 $\pm$ 7 & 0.044 $\pm$ 0.049 \\
\cmidrule{2-8}
& Fwd Loop & 4.5 $\pm$ 5.2 & 3.0 $\pm$ 3.9 & 34.3 $\pm$ 31.1 & 1.6k $\pm$ 2.1k & 8 $\pm$ 11 & 0.056 $\pm$ 0.066 \\
& Bwd Loop & 5.6 $\pm$ 6.0 & 3.3 $\pm$ 4.2 & 37.1 $\pm$ 31.9 & 2.4k $\pm$ 3.3k & 3 $\pm$ 3 & 0.174 $\pm$ 0.169 \\
\cmidrule{2-8}
& RR & \textbf{14.3 $\pm$ 8.5} & \textbf{4.2 $\pm$ 4.8} & 70.1 $\pm$ 24.4 & 2.4k $\pm$ 3.2k & 18 $\pm$ 16 & 0.088 $\pm$ 0.125 \\
& DTS & 14.0 $\pm$ 8.4 & 4.1 $\pm$ 4.7 & \textbf{72.0 $\pm$ 23.5} & 2.5k $\pm$ 3.3k & 17 $\pm$ 16 & 0.089 $\pm$ 0.105 \\
\midrule
\multirow{8}{*}{RRT} & Composite & 17.7 $\pm$ 18.2 & 17.5 $\pm$ 18.2 & 7.9 $\pm$ 30.9 & 2.9k $\pm$ 3.8k & 5 $\pm$ 4 & 0.359 $\pm$ 0.278 \\
& Prioritized & 22.7 $\pm$ 22.0 & 23.6 $\pm$ 22.5 & 6.2 $\pm$ 46.0 & 2.3k $\pm$ 2.8k & 17 $\pm$ 15 & 0.129 $\pm$ 0.145 \\
& Path & 23.4 $\pm$ 20.7 & 26.0 $\pm$ 22.1 & \textbf{46.0 $\pm$ 35.9} & 2.0k $\pm$ 2.4k & 48 $\pm$ 44 & 0.063 $\pm$ 0.047 \\
& TPG & 9.4 $\pm$ 13.7 & 12.3 $\pm$ 14.7 & -6.5 $\pm$ 31.2 & 0.6k $\pm$ 1.1k & 12 $\pm$ 6 & 0.064 $\pm$ 0.087 \\
\cmidrule{2-8}
& Fwd Loop & 14.7 $\pm$ 16.8 & 18.0 $\pm$ 18.2 & 23.1 $\pm$ 34.5 & 1.1k $\pm$ 1.6k & 54 $\pm$ 55 & 0.034 $\pm$ 0.029 \\
& Bwd Loop & 18.0 $\pm$ 19.1 & 21.8 $\pm$ 20.7 & 33.1 $\pm$ 41.8 & 2.3k $\pm$ 3.2k & 6 $\pm$ 6 & 0.472 $\pm$ 0.578 \\
\cmidrule{2-8}
& RR & \textbf{25.1 $\pm$ 21.8} & \textbf{27.8 $\pm$ 23.2} & 33.7 $\pm$ 46.4 & 3.0k $\pm$ 3.5k & 46 $\pm$ 39 & 0.055 $\pm$ 0.057 \\
& DTS & 25.1 $\pm$ 21.7 & 27.6 $\pm$ 23.0 & 36.1 $\pm$ 39.8 & 3.1k $\pm$ 3.6k & 44 $\pm$ 38 & 0.060 $\pm$ 0.066 \\
\bottomrule
\end{tabular}
    \vspace{-0.4cm}
\end{table*}

\begin{figure*}[t]
    \centering
    \includegraphics[width=0.4\textwidth]{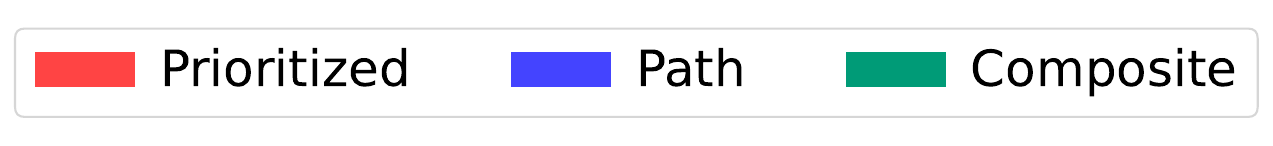}

    \begin{subfigure}[b]{0.49\textwidth}
        \centering
        \includegraphics[width=\columnwidth]{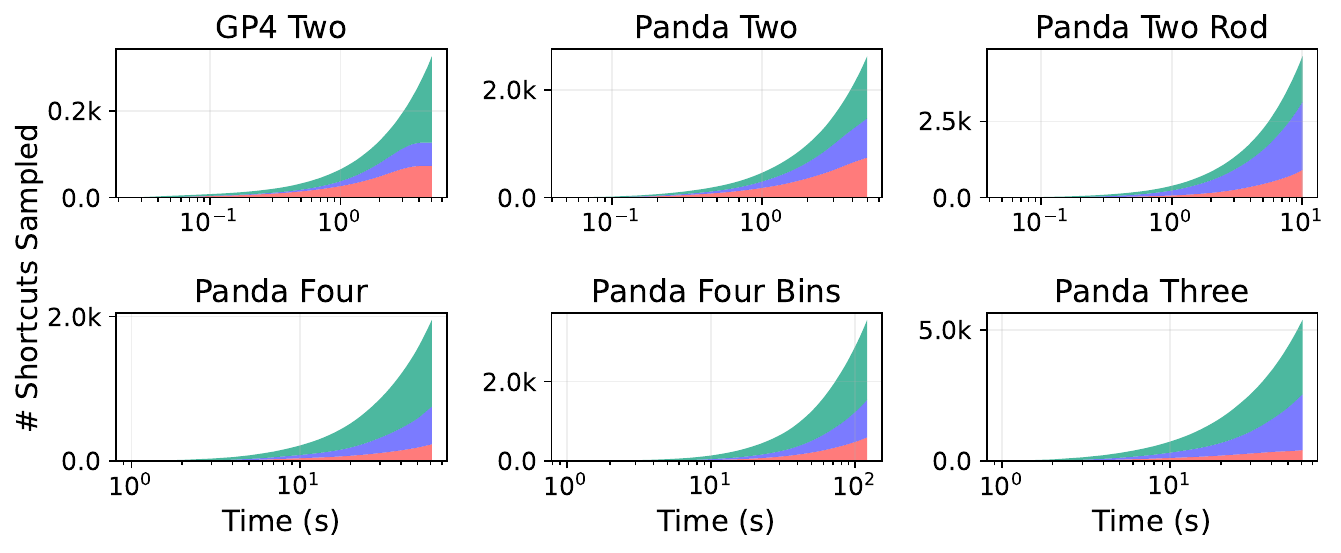}
        \caption{Shortcut Sampled}
        \label{fig:cbs_thompson_ selection}
    \end{subfigure}
    \hfill
    \begin{subfigure}[b]{0.49\textwidth}
        \centering
        \includegraphics[width=\columnwidth]{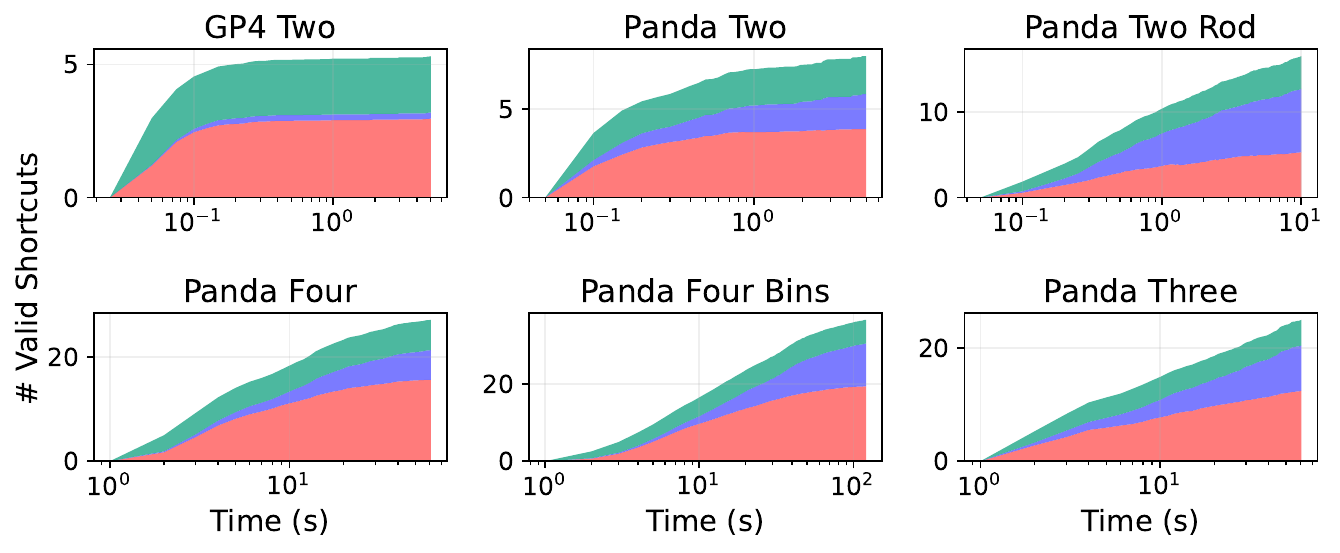}
        \caption{Shortcut Valid}
        \label{fig:cbs_thompson_valid_selection}
    \end{subfigure}
    
    \caption{Distribution of shortcut sampled and valid shortcuts found by each underlying method (Composite, Path, Prioritized) of the DTS sampling strategy. The plots show the average accumulated number of valid shortcuts by each underlying method across all CBS-generated trajectories in each environment.}
    \label{fig:selection_cbs_thompson}
    \vspace{-0.6cm}
\end{figure*}

\subsubsection{Single-Robot Shortcutting Methods}
To compare how single-robot shortcutting performs, we examine Path, Fwd Loop, and Bwd Loop, all of which use path shortcutting but differ in how shortcut endpoints are sampled. In Table \ref{tab:agg_stat}, the randomized strategy (Path) dominates both iterative strategies in makespan, path length, and smoothness. Fig. \ref{fig:makespan_comparison} also shows that the randomized strategy converges significantly faster. The additional statistics support earlier analysis in Sec. \ref{sec:single-robot-comp}. Compared to randomized sampling, forward iteration has a smaller makespan improvement per valid shortcut, while backward iteration finds fewer valid shortcuts.
    
\subsubsection{Multi-Robot Shortcutting Methods}

From the anytime makespan-runtime curve in Fig. \ref{fig:makespan_comparison} and final performance in Table \ref{tab:agg_stat}, composite shortcutting reduces the makespan faster than prioritized and path shortcutting, but converges to less optimal trajectories. Prioritized shortcutting achieves an overall makespan improvement on par or better than path shortcutting, but path shortcutting converges faster than prioritized shortcutting for most RRT-generated trajectories. Path shortcutting reduces the final path length more than composite or prioritized shortcutting methods. 
The statistics in Table \ref{tab:agg_stat} reveal an important detail. Composite shortcutting is much more greedy than prioritized or path shortcutting, as it finds fewer valid shortcuts but with greater makespan improvement per shortcut. Lastly, TPG shortcutting converges more slowly to worse final trajectories, and also evaluates less candidate shortcuts as each evaluation is more expensive.

\subsubsection{Multi-Strategy Method}
From the runtime-performance curves in Fig. \ref{fig:makespan_comparison}, the DTS method dominates all other methods throughout the runtime curve. DTS is as efficient as composite shortcutting early on and still converges to high-quality final trajectories compared to path or prioritized shortcutting. The round-robin (RR) method is less efficient but still converges to equally good trajectories in terms of makespan. DTS also generates smoother final trajectories than prioritized or composite shortcutting, except against path shortcutting for RRT-generated trajectories.

\begin{figure}[tbp!]
    \centering
    \includegraphics[width=1.0\linewidth]{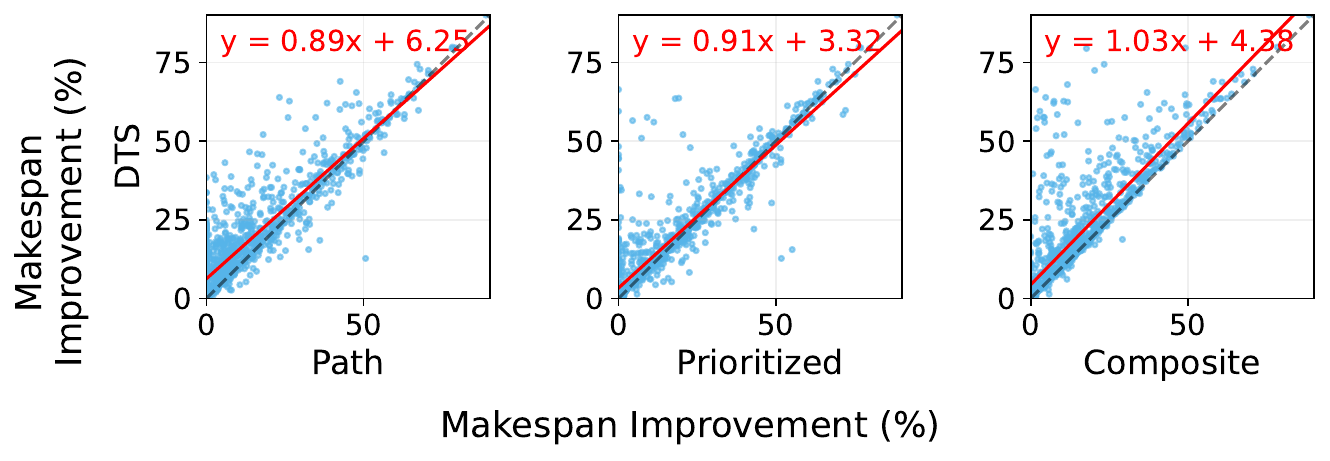}
    \caption{Makespan improvement for all trajectories generated by both planners in all six environments. 
    } 
    \label{fig:multi_scatter}
    \vspace{-0.8cm}
\end{figure}

To help better illustrate the distribution of the final makespan between DTS and underlying methods beyond just the mean and standard deviation, we compare the results of all 1,034 trajectories in Fig. \ref{fig:multi_scatter}. Each dot represents the final makespan improvement of DTS on the x-axis and one randomized shortcutting method on the y-axis. DTS almost always outperforms the composite shortcutting methods. The performance gap is not as significant for path and prioritized shortcuts, but still skews toward DTS, with many more outperforming cases than underperforming ones. 

Lastly, we analyze how the multi-strategy method performs better than its components. As shown in Table \ref{tab:agg_stat}, the average number of valid shortcuts found by multi-strategy methods exceeds that by underlying methods. The distribution of shortcuts sampled (Fig. \ref{fig:cbs_thompson_ selection}) and valid shortcuts (Fig. \ref{fig:cbs_thompson_valid_selection}) found by each method over time is often inconsistent depending on the domain. Also, DTS prefers path and composite shortcutting over prioritized shortcutting due to the prior and shorter runtime. These observations suggest that DTS can effectively adapt to different domains and converge quickly to high-quality final trajectories. 

\vspace{-0.2cm}
\subsection{Real Robot Experiment}
\vspace{-0.1cm}
We conduct real experiments in the Dual GP4 environment, qualitatively comparing RRT-generated trajectories before and after 0.1-second DTS shortcutting. With shortcutting, the robot motions are still collision-free in a more physically confined space, avoiding unnecessary movements. Shortcutting reduces makespan by 20\% to 58\% in the five examples tested. The results are available in our \href{https://philip-huang.github.io/mr-shortcut/}{video}.

\section{Conclusion}
In this work, we examine an overlooked but critical problem in the literature for M-RAMP -- \textit{shortcutting}. To the best of our knowledge, this is the first work that analyzes existing shortcutting algorithms for M-RAMP. Most importantly, shortcutting can significantly improve the makespan by 14\% and 25\% on average for 1034 CBS- and RRT-generated trajectories, respectively. We find that randomly sampling shortcuts performs better compared to deterministic iteration. Comparing the four multi-robot shortcutting strategies, we show that composite shortcutting is more efficient but converges to less optimal trajectories. Prioritized and path shortcutting converge to final trajectories with shorter makespan and smoother paths, but the better-performing method often depends on the trajectory. TPG shortcutting is very computationally expensive. We then present two simple strategies that adaptively combine the strengths of these strategies. Our multi-strategy method achieves stronger anytime performance and can be used on a real dual GP4 environment. An exciting future work is shortcutting, long-horizon, multi-task trajectories for multi-arm manipulation.

\section*{Acknowledgments}
This work is in part supported by the National Science Foundation (NSF) under grant number 2328671 and the Manufacturing Futures Institute, Carnegie Mellon University, through a grant from the Richard King Mellon Foundation, as well as a gift from Amazon. 

\bibliographystyle{IEEEtran}
\bibliography{ref_abbreviated}

\clearpage
\appendix

\vspace{0.3cm}
\subsection{Details of Thompson Sampling}
\vspace{-0.2cm}

\begin{table}[htbp]
\centering
\caption{Thompson Sampling Parameters for Dynamic Shortcut Selection}
\label{tab:thompson_params}
\begin{tabular}{l|c|l}
\hline
\textbf{Parameter} & \textbf{Value} & \textbf{Description} \\
\hline
\multicolumn{3}{l}{\textbf{Time and Reward Parameters}} \\
\hline
$\sigma$ & 0.01 & Normalizing time scale (second) \\
$\gamma_t$  & 1.0 & Relative weight to reward shorter runtime \\
\hline
\multicolumn{3}{l}{\textbf{Beta Distribution Parameters by Strategy}} \\
\hline
$\alpha_{\text{composite}}$ & 10 & Initial alpha for composite shortcutting \\
$\beta_{\text{composite}}$ & 1 & Initial beta for composite shortcutting \\
\hline
$\alpha_{\text{prioritized}}$ & 1 & Initial alpha for prioritized shortcutting \\
$\beta_{\text{prioritized}}$ & 1 & Initial beta for prioritized shortcutting \\
\hline
$\alpha_{\text{path}}$ & 1 & Initial alpha for path shortcutting \\
$\beta_{\text{path}}$ & 1 & Initial beta for path shortcutting \\
\hline
\multicolumn{3}{l}{\textbf{Update Parameters}} \\
\hline
$Z$ & 1000 & Normalization constant \\
$\gamma_\alpha$ & 100.0 & Update weight for $\alpha$ parameters \\
$\gamma_\beta$ & 0.1 & Update weight for $\beta$ parameters \\
\hline
\end{tabular}
\end{table}

These parameters were chosen to optimize the performance of our Thompson Sampling multi-strategy shortcutting method. By setting $\alpha_{\text{composite}}= 10$ in the prior distribution, this increases the likelihood of choosing a composite shortcut early on and leads to improved convergence speed similar to that of the composite shortcut. This is an advantage of the Thompson method compared to the Round-Robin method, which converges to similar results but often at a slower rate.

Algorithm \ref{alg:dts} shows the pseudocode of the Thompson sampling multi-strategy shortcutting method.

\begin{algorithm}[h!]
\caption{Dynamic Thompson Sampling}
\label{alg:dts}
\begin{algorithmic}[1]
\State{Initialize $\alpha_i$ and $\beta_i$ for each strategy}
\While{time limit not exceeded}
    \For{each strategy $i$}
        \State{Sample $\theta_i \sim \text{Beta}(\alpha_i, \beta_i)$}
    \EndFor
    \State{Select strategy $i^* = \argmax_{i} \theta_i$ }
    \State{Run shortcut strategy $i^*$ and get reward $r$}
    \State{$\alpha_{i^*} \gets \alpha_{i^*} + \gamma_{\alpha} \cdot r$ if shortcut is valid}
    \State{$\beta_{i^*} \gets \beta_{i^*} + \gamma_{\beta}$ if shortcut is invalid}
    \If{$(\alpha_{i^*} + \beta_{i^*}) > Z$}
        \State{$\alpha_{i^*} \gets \frac{\alpha_{i^*}}{\alpha_{i^*} + \beta_{i^*}} \cdot Z$}
        \State{$\beta_{i^*} \gets \frac{\beta_{i^*}}{\alpha_{i^*} + \beta_{i^*}} \cdot Z$}
    \EndIf
\EndWhile
\end{algorithmic}
\vspace{-0.1cm}
\end{algorithm}

\subsection{Details of Real Robot Setup}
To enable the generated trajectories for real robot execution, we require a single-robot controller as well as a synchronization method to coordinate the execution of all robot arms according to the planned multi-robot trajectory. We follow the TPG-based execution pipeline in APEX-MR \cite{huang2025apexmr}. We use the position controller, and each robot has its own position controller that runs in a separate process. We build a TPG based on the multi-robot trajectory and host the TPG on a centralized server. During execution, the TPG sends actions that are safe to be executed to each robot's action queue. Each robot executes all the enqueued actions and updates the status back to the central TPG server. The server then repeatedly enqueues new nodes that become safe to each robot's action queue. Using TPG ensures no robot-robot collision even in the case of controller delay and allows the aggressive post-processed trajectories to be safely executed.

\end{document}